\documentclass{article} 
\pdfoutput=1
\usepackage{nips13submit_e,times}
\usepackage{url}
\usepackage{graphicx} 
\usepackage{caption}
\usepackage{subcaption}
\usepackage{amsmath,amsfonts,amsthm,amssymb}
\usepackage{authblk}
\usepackage{booktabs}

\title{Learning unbiased features}

\author[1]{\textbf{Yujia Li}}
\author[1]{\textbf{Kevin Swersky}}
\author[1,2]{\textbf{Richard Zemel}}
\affil[1]{University of Toronto}
\affil[2]{Canadian Institute for Advanced Research}
\affil[ ]{\texttt{\{yujiali, kswersky, zemel\}@cs.toronto.edu}}

\nipsfinalcopy 

\begin{document}

\maketitle


\section{Introduction}
A key element in transfer learning is representation learning;
if representations can be developed that expose the
relevant factors underlying the data, then
new tasks and domains can be learned readily based on mappings
of these salient factors.
We propose that an important aim for these representations are
to be {\em unbiased}.  
Different forms of representation learning can be derived
from alternative definitions of unwanted bias, e.g.,
bias to particular tasks, domains, or irrelevant underlying data dimensions.
One very useful approach to estimating the amount of bias in
a representation comes from
maximum mean discrepancy (MMD) \cite{gretton2006kernel}, a measure of 
distance between probability distributions.
We are not the first to suggest that MMD can be a useful
criterion in developing representations that apply across multiple
domains or tasks \cite{baktashmotlagh2013unsupervised}.  However,
in this paper we describe a number of novel applications of this criterion
that we have devised, all based on the idea of developing unbiased 
representations.
These formulations include: a standard domain adaptation framework; a
method of learning invariant representations; an approach based on
noise-insensitive autoencoders; and a novel form of generative model.
We suggest that these formulations are relevant for the transfer
learning workshop for a few reasons:
(a). they focus on deep learning;
(b). the formulations include both supervised and unsupervised learning 
scenarios; and
(c). they are well-suited to the scenario emphasized in the call-for-papers,
where the learning task is not focused on the regime of limited training data
but instead must manage large scale data, which may be limited in labels
and quality.

\section{Maximum Mean Discrepancy}
Each of our approaches to learn unbiased features rely on a
sample-based measure of the bias in the representation.
A two sample test is a statistical test that tries to determine, given two
datasets $\{X_n\}\sim P$ and $\{Y_m\}\sim Q$, whether the datasets have been generated from the same underlying distribution, i.e., if $P=Q$. Maximum mean discrepancy~\cite{gretton2006kernel} is a useful distance measure between two distributions that can be used to perform two sample tests.
\small
\begin{align}
\mathrm{MMD}(X,Y) =& \left\|\frac{1}{N}\sum_{n=1}^{N} \phi(X_n) -
    \frac{1}{M}\sum_{m=1}^{M} \phi(Y_m)\right\|^2 \\
=& \frac{1}{N^2}\sum_{n=1}^{N} \sum_{n'=1}^{N} \phi(X_n)^\top\phi(X_{n'}) + \frac{1}{M^2}\sum_{m=1}^{M} \sum_{m'=1}^{M} \phi(Y_m)^\top\phi(Y_{m'}) - \frac{2}{NM}\sum_{n=1}^{N}\sum_{m=1}^{M} \phi(X_n)^\top\phi(Y_m) \label{eq:mmd_expanded}
\end{align}
\normalsize
Where $\phi(\cdot)$ is a feature expansion function. We can apply the kernel
trick to each inner product in Equation (\ref{eq:mmd_expanded}) to use an
implicit feature space. When the space defined by the kernel is a universal
reproducing kernel Hilbert space then asymptotically MMD is $0$ if and only if $P=Q$~\cite{gretton2012kernel}.

\section{Applications}
\subsection{Domain Adaptation}
In domain adaptation, we are given a set of labeled data from a source domain and a set of
unlabeled data from a different target domain. The task is to learn a model
that works well on the target domain.

In our framework, we want to learn unbiased features that are invariant to the nuances
across different domains.  The classifier trained on these features can then
generalize well over all domains. We use deep neural networks as the
classification model. MMD is used as a penalty on one hidden layer of the neural net to
drive the distributions of features for the source and target domains to be close to each
other.  While the use of MMD is similar to that of
\cite{baktashmotlagh2013unsupervised}, we use a neural network to learn both
the features and classifier jointly. The distributed representation of a
neural network is far more powerful than the linear transformation and clustering method proposed in
\cite{baktashmotlagh2013unsupervised}.

We tested the neural network with MMD penalty model on the Amazon product
review sentiment classification dataset \cite{blitzer2007biographies}. This
dataset contains product reviews from 4 domains corresponding to 4 product
categories (books, dvd, electronics, kitchen).  Each review is labeled either
positive or negative, and we preprocessed them as TF-IDF vectors. 
We tested a 2 hidden layer neural net model on the adaptation tasks between
all pairs of source and target domains. For each task, a small portion of the
the labeled source domain data is used as validation data for early
stopping. Other hyper parameters are chosen to optimize the average target performance over
10 random splits of the data, in a setting similar to cross-validation. The
best target accuracy with standard deviation for a few tasks are shown
in Table \ref{tbl:da}. More results and experiment settings can be found in the
appendix.

We compare our method with SVM models with no adaptation, neural net with the
same architecture but no MMD penalty, and another popular domain adaptation
baseline Transfer Component Analysis (TCA) \cite{pan2011domain}. The neural
net model with MMD penalty dominates on most tasks. Even with the more basic 
word count features the ``NN MMD'' method still works better than most other
baselines, demonstrating the ability of our model to learn features useful
across domains.

\begin{table}[t]
    \centering
    \begin{tabular}{r|c|c|c|c|c|c}
        \toprule
                  &  E$\rightarrow$B   &
             B$\rightarrow$D  &  K$\rightarrow$D  &
                 D$\rightarrow$E  &  B$\rightarrow$K  &
             E$\rightarrow$K  \\
        \hline
        \hline
        Linear SVM & 71.0 $\pm$ 2.0 & 79.0
         $\pm$ 1.9 & 73.6 $\pm$ 1.5 & 74.2
         $\pm$ 1.4 & 75.9 $\pm$ 1.8 & 84.5
        $\pm$ 1.0  \\
        \hline
        RBF SVM & 68.0 $\pm$ 1.9 & 79.1
      $\pm$ 2.3 & 73.0 $\pm$ 1.6 & 76.3
      $\pm$ 2.2 & 75.8 $\pm$ 2.1 & 82.0
        $\pm$ 1.4  \\
        \hline
        TCA & 71.8 $\pm$ 1.4 & 76.9 $\pm$
        1.4 & 73.3 $\pm$ 2.4 & 75.9 $\pm$
        2.7 & 76.8 $\pm$ 2.1 & 80.2 $\pm$
        1.4  \\
        \hline
        NN & 70.0 $\pm$ 2.4 & 78.3 $\pm$ 1.6
           & 72.7 $\pm$ 1.6 & 72.8 $\pm$ 2.4
           & 74.1 $\pm$ 1.6 & 84.0 $\pm$ 1.5
        \\
        \hline
        \hline
        NN MMD${}^*$ & 71.8 $\pm$ 2.1 & 77.4 $\pm$
           2.4 & 73.9 $\pm$ 2.4 & 78.4 $\pm$
           1.6 & 77.9 $\pm$ 1.6 & 84.7 $\pm$
        1.6  \\
        \hline
        NN MMD & \textbf{73.7 $\pm$ 2.0} & \textbf{79.2 $\pm$
          1.7} & \textbf{75.0 $\pm$ 1.0} & \textbf{79.1 $\pm$
          1.6} &\textbf{78.3 $\pm$ 1.4} & \textbf{85.2 $\pm$
          1.1}  \\
        \bottomrule
    \end{tabular}
    \vspace{-5pt}
    \caption{Domain adaptation results for product review sentiment
    classification task. NN MMD${}^*$: neural net with MMD trained and tested on word count
    instead of TF-IDF features.}
    \label{tbl:da}
    \vspace{-5pt}
\end{table}

\subsection{Learning Invariant Features}
In this application we use the proposed framework to learn
features invariant to transformations on input data. More specifically, we
want to learn features for human faces that are both good for identity
recognition and invariant to different lighting conditions.

In the experiment we used the extended Yale B dataset, which contains faces of
38 people under various lighting conditions corresponding to light source from
different directions. We created $5$ groups of images, corresponding to light
source in upper right, lower right, lower left, upper left and the front. Then
for each group of images, we chose $1$ image per person to form one domain for
that lighting condition. In this way we had $5$ domains with $5 \times 38 = 190$ images
in total. All the other images (around $2000$) are used for testing. The 
task is to recognize the identity of the person in image, i.e. a $38$-way
classification task. For this task, we did not use a validation set, but rather
report the best result on test set to see where the limits of different models
are. Note that the lighting conditions here can be modeled very well with a Lambertian
model, however we did not use this strong model but rather choose to use a generic
neural network to learn invariant features, so that the proposed method can
be readily applied to other applications. 

The proposed model for this task is similar to the one used in the previous
section, except that the MMD penalty is now applied to the distribution of
hidden representations for $5$ different domains rather than two. We used the following formulation which is a sum of MMD between
each individual distribution and the average distribution across all domains
\begin{equation}
\mathrm{MMD} = \sum_{s=1}^S \left\|\frac{1}{N_s} \sum_{i:d_i=s} \phi(h_i)
- \frac{1}{N} \sum_n \phi(h_n)\right\|^2
\end{equation}
where $s$ indexes domains, $i$ indexes examples, $S=5$ is the number of different domains, $N_s$ is the
number of examples from domain $s$, $N$ is the total number of examples across
all domains, $d_i$ is the domain label for example $i$ and $h_i$ is the hidden representation computed
from a neural network. We use a two hidden layer neural net with $256$ and $128$ ReLU units on each of
them for this task. The MMD penalty with a Gaussian kernel is applied to the second hidden layer.
Dropout~\cite{hinton2012improving} is used for all the methods compared here to regularize the network as overfitting
is a big problem.

On this task, the baseline model trained without the MMD penalty achieves a test accuracy of
72\% (100\% training accuracy). Using the MMD penalty with Gaussian kernel, the best test accuracy
improved significantly to around 82\%. Using a linear kernel leads to a test accuracy to 78\%.

\begin{figure}[t]
    \centering
    \begin{tabular}{c|c}
        \includegraphics[width=0.47\textwidth]{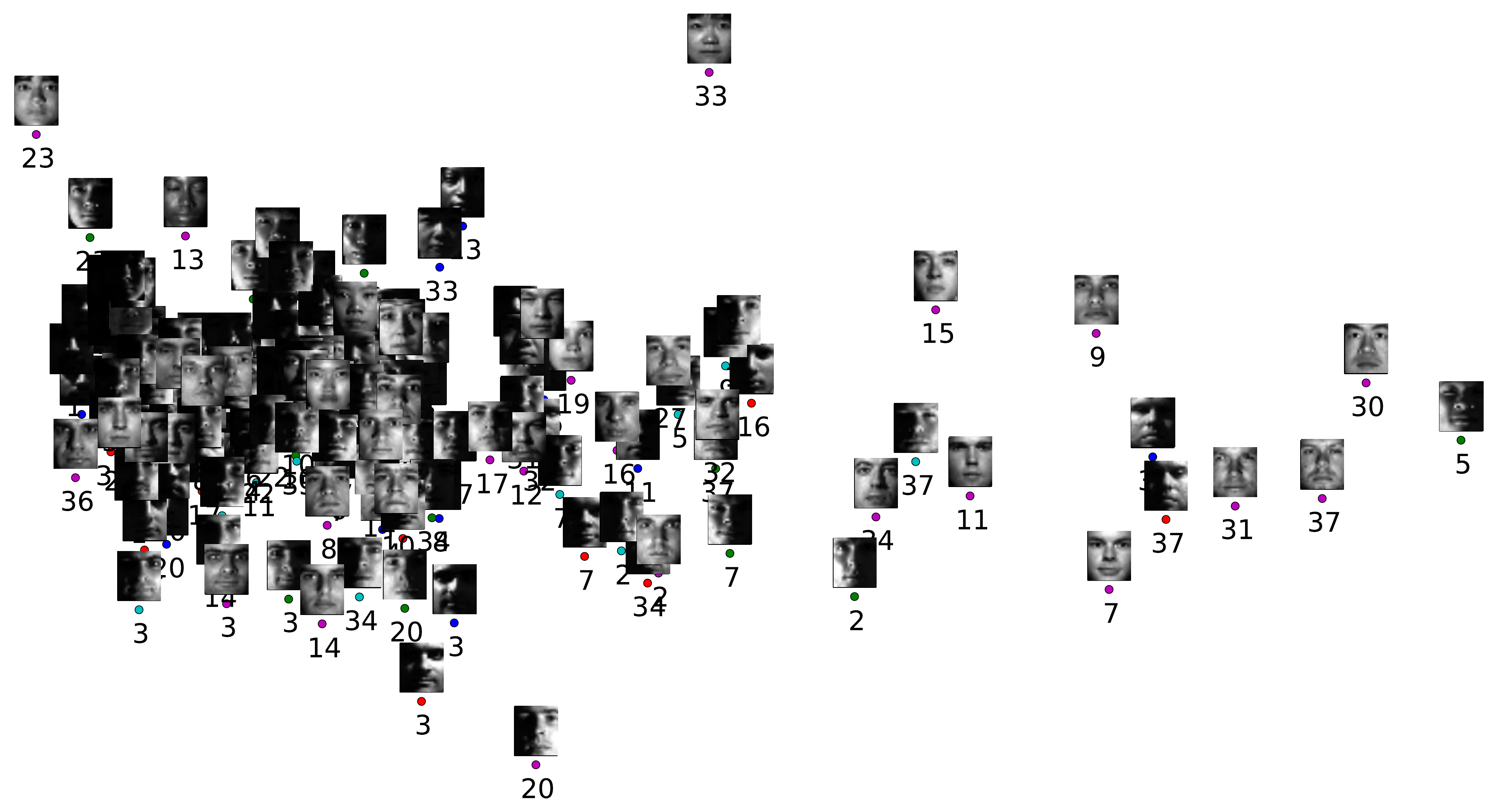} &
        \includegraphics[width=0.47\textwidth]{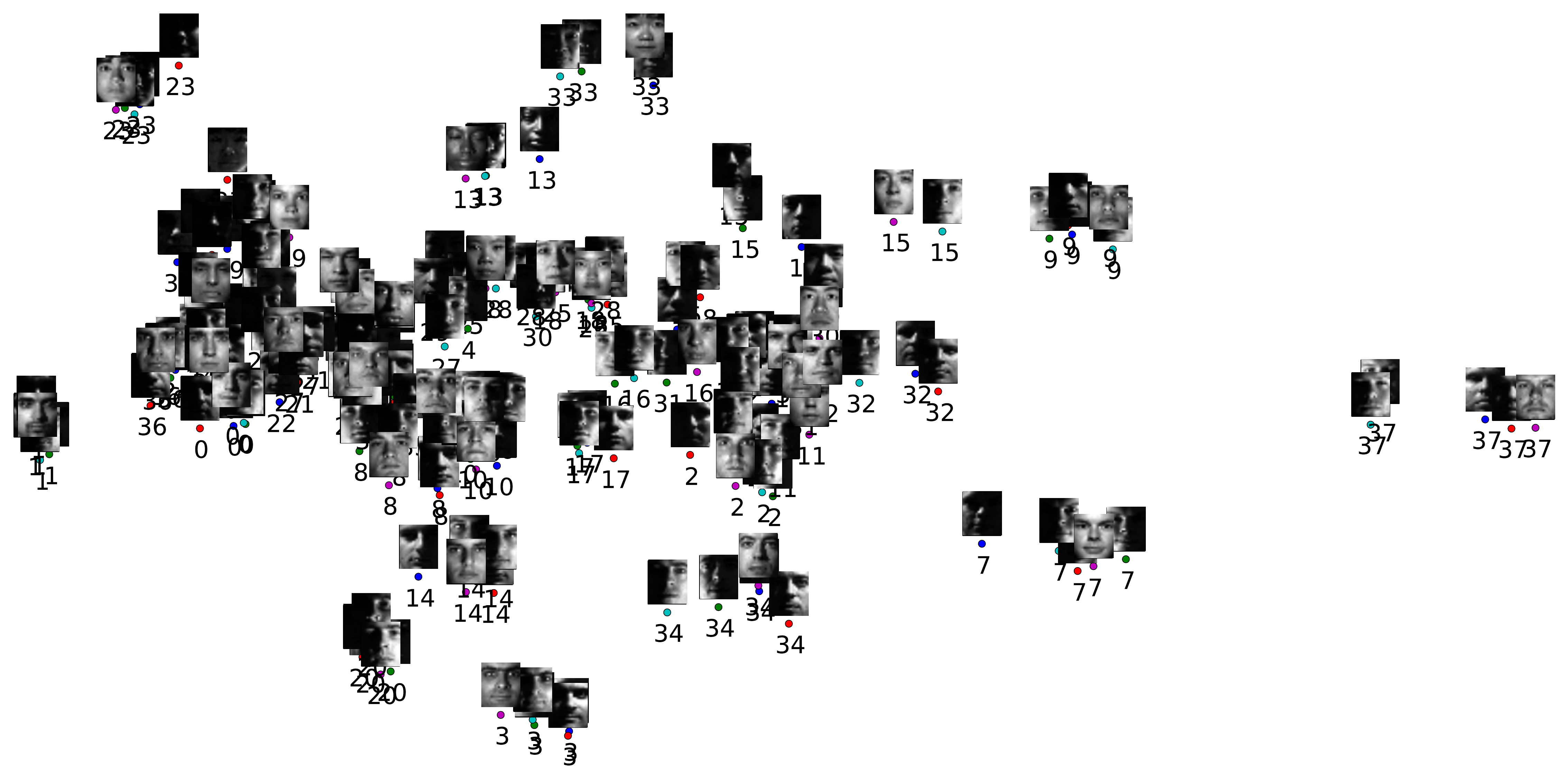} \\
        (a) Without MMD & (b) With MMD
    \end{tabular}
    \caption{PCA projection of the representations of the second hidden layer
    for the training images. Each example is plotted with the person ID and
    the image. Zoom in to see the details.}
    \label{fig:mmdface}
\end{figure}

We visualize the hidden representations for the training images learned with the
Gaussian kernel MMD penalty in Figure \ref{fig:mmdface}.
Note that examples for each person
under very different lighting conditions are
grouped together even though the MMD penalty only depends on lighting condition, and does not take into account identity.

\subsection{Noise-Insensitive Autoencoders}
Auto-encoders (AEs) are neural network models that have two basic components: an encoder, that maps data into a latent space, and a decoder, that maps the latent space back out into the original space. Auto-encoders are typically trained to minimize reconstruction loss from encoding and decoding. In many applications, reconstruction loss is merely a proxy and can lead to spurious representations. Researchers have spent a great deal of effort developing new regularization schemes to improve the learned representation~\cite{swersky2011autoencoders, vincent2010stacked, rifai2011contractive}. Two such methods include denoising auto-encoders (DAEs)~\cite{vincent2010stacked} and contractive auto-encoders (CAEs)~\cite{rifai2011contractive}. With denoising auto-encoders, the data is perturbed with noise and the reconstruction loss is altered to measure how faithfully the original data can be recovered from the pertrubed data. Contractive auto-encoders more explicitly penalize the latent representation so that it becomes invariant to infinitesimal perturbations in the original space. In the appendix, we show how the CAE penalty can be interpreted as a form of MMD penalty with a linear kernel.

We experiment with several single-layer auto-encoder variants, including an ordinary auto-encoder trained on reconstruction loss, a contractive auto-encoder, and a denoising auto-encoder. For comparison, we augment both the ordinary auto-encoder and denoising auto-encoder with the MMD penalty on their hidden layer, sampling a new set of perturbed hidden units with each weight update. We trained each model on $10,000$ MNIST digits and tuned hyperparameters to minimize a denoising reconstruction loss on held-out data. Further details can be found in the appendix.

To measure the invariance to perturbation, we created a noisy copy of the test data and trained an SVM classifier on the latent representations to distinguish between clean and noisy data. A worse accuracy corresponds to a more unbiased latent representation. The MMD autoencoder outperformed the other approaches on this measure. Surprisingly, the denoising autoencoder performed the worst, demonstrating that denoising does not necessarily produce features that are invariant to noise. Also interesting is that a relatively low contraction penalty was chosen for the CAE, as higher penalties seemed to incur higher denoising reconstruction loss. This is likely due to the difference between the applied Bernoulli noise, and the infintesimal noise assumed by the CAE. Plots of the filters can be found in the appendix.

\begin{table}[h]
\centering
\begin{tabular}{|r|c|c|c|c|c|}
\hline
Model        & AE   & DAE  & CAE  & MMD  & MMD+DAE \\ \hline\hline
SVM Accuracy & 78.6 & 82.5 & 77.9 & 61.1 & 72.9    \\ \hline
\end{tabular}
\caption{SVM accuracy on distinguishing clean from noisy data. Lower accuracy means the learned features are more invariant to noise.}
\end{table}

\subsection{Learning Generative Deep Models}
The last application we consider is to use the MMD criterion for learning
generative models. Unlike previous sections where MMD is used to learn
unbiased representations, in this application we use MMD to match the 
distribution of the generative model with the data distribution.  The idea is
MMD should be small on samples from a good generative model.

Here we train a generative deep model proposed in
\cite{goodfellow2014generative} on a subset of 1000 MNIST digits. The model
contains a stochastic hidden layer $h$ at the top with a fixed prior distribution
$p(h)$, and a mapping $f$ that deterministically maps $h$ to $x$. The prior
$p(h)$ and the mapping $f(x|h)$ together implicitly defines the distribution
$p(x)$.

\begin{figure}[t]
    \centering
    \begin{tabular}{cc}
        \includegraphics[width=0.4\textwidth]{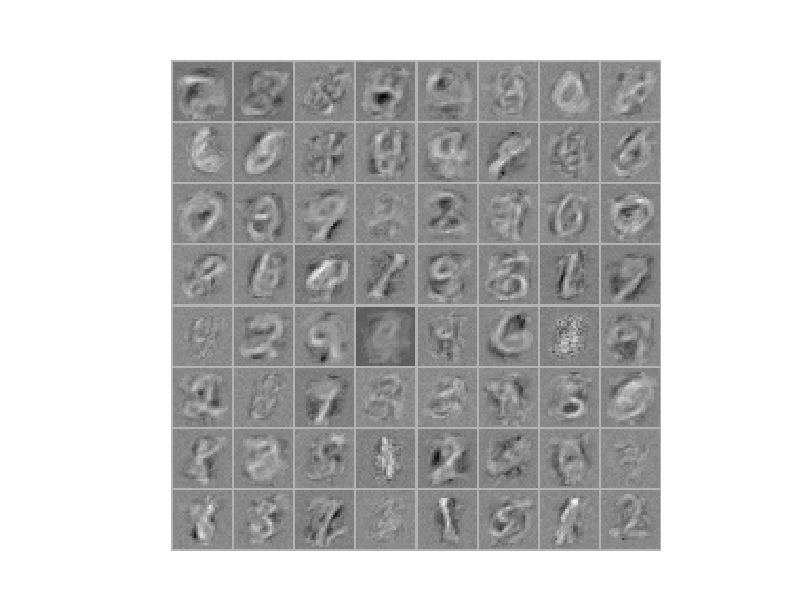} & 
        \includegraphics[width=0.4\textwidth]{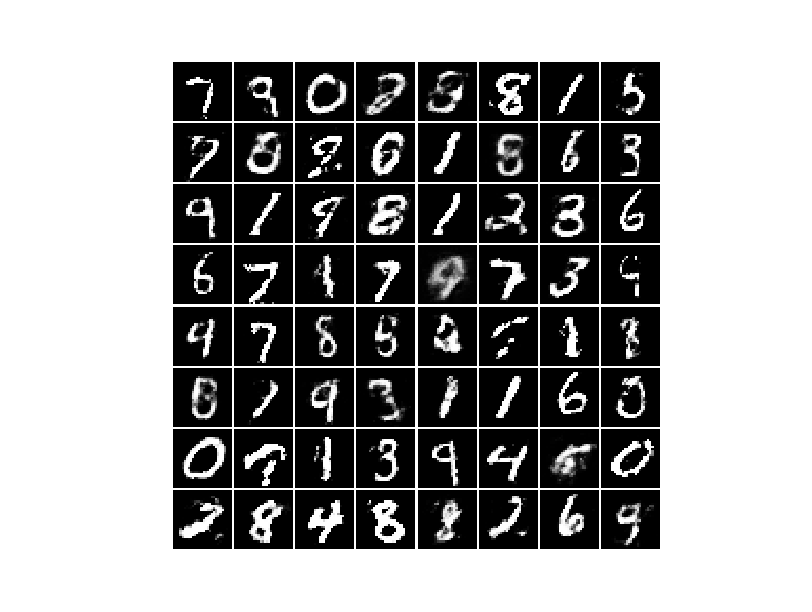} \\
        (a) & (b)
    \end{tabular}
    \caption{(a) visualization of some bottom layer weights; (b) independent
    samples from the model.}
    \label{fig:gen}
\end{figure}

In \cite{goodfellow2014generative} the authors proposed a minimax formulation
to learn the mapping $f$, where one extra classifier looks at the data and the samples of the
model and then try to do a good job of distinguishing them, and the parameters
of $f$ is updated to make this classifier do as bad as possible so that
samples generated will be close to the data. As the formulation interleaves
two optimization problems with opposite objectives, careful scheduling is
required for the model to converge to a good point.

We propose to directly minimize the MMD between the data and the model
samples. Given a fixed sample of $h$, we can backpropagate through the MMD
penalty and the whole network, to drive the model samples to be close to the
data. This method utilizes a single consistent
objective and completely avoids the minimax problem. Details of our architecture and training can be found in the appendix.

Figure \ref{fig:gen} visualizes some bottom layer weights of the network and a
set of samples generated from the model. We can see that with this method the
model learns some meaningful features and is able to generate realistic
samples.


{\small
\bibliography{unbiasedFeatures-workshop}
\bibliographystyle{plain}
}
\clearpage

\section{Appendix}
\subsection{More Details of the Domain Adaptation Experiments}

\begin{table}[t]
    \centering
    \begin{tabular}{r|c|c|c|c|c|c}
        \toprule
                 & D$\rightarrow$B  &  E$\rightarrow$B  &  K$\rightarrow$B  &
             B$\rightarrow$D  &  E$\rightarrow$D  &  K$\rightarrow$D \\
        \hline
        \hline
        Linear SVM & 78.3 $\pm$ 1.4 & 71.0 $\pm$ 2.0 & 72.9 $\pm$ 2.4 & 79.0
         $\pm$ 1.9 & 72.5 $\pm$ 2.9 & 73.6 $\pm$ 1.5 \\
        \hline
        RBF SVM & 77.7 $\pm$ 1.2 & 68.0 $\pm$ 1.9 & 73.2 $\pm$ 2.4 & 79.1
      $\pm$ 2.3 & 70.7 $\pm$ 1.8 & 73.0 $\pm$ 1.6 \\
        \hline
        TCA & 77.5 $\pm$ 1.3 & 71.8 $\pm$ 1.4 & 68.8 $\pm$ 2.4 & 76.9 $\pm$
        1.4 & 72.5 $\pm$ 1.9 & 73.3 $\pm$ 2.4  \\
        \hline
        NN & 76.6 $\pm$ 1.8 & 70.0 $\pm$ 2.4 & 72.8 $\pm$ 1.5 & 78.3 $\pm$ 1.6
           & 71.7 $\pm$ 2.7 & 72.7 $\pm$ 1.6
        \\
        \hline
        NN MMD${}^*$ & 76.5 $\pm$ 2.5 & 71.8 $\pm$ 2.1 & 72.8 $\pm$ 2.4 & 77.4 $\pm$
           2.4 & 74.3 $\pm$ 1.7 & 73.9 $\pm$ 2.4  \\
        \hline
        NN MMD & \textbf{78.5 $\pm$ 1.5} & \textbf{73.7 $\pm$ 2.0} &
        \textbf{75.7 $\pm$ 2.3} & \textbf{79.2 $\pm$ 1.7} & \textbf{75.3 $\pm$
                           2.1} & \textbf{75.0 $\pm$ 1.0} \\
        \hline\hline
                & B$\rightarrow$E
                 &  D$\rightarrow$E  &  K$\rightarrow$E  &  B$\rightarrow$K  &
             D$\rightarrow$K  &  E$\rightarrow$K  \\
        \hline
        \hline
        Linear SVM & 72.4 $\pm$ 3.0 & 74.2
         $\pm$ 1.4 & 82.7 $\pm$ 1.3 & 75.9 $\pm$ 1.8 & 77.0 $\pm$ 1.8 & 84.5
        $\pm$ 1.0  \\
        \hline
        RBF SVM & 72.8 $\pm$ 2.5 & 76.3
      $\pm$ 2.2 & 82.5 $\pm$ 1.4 & 75.8 $\pm$ 2.1 & 76.0 $\pm$ 2.2 & 82.0
        $\pm$ 1.4  \\
        \hline
        TCA & 72.1 $\pm$ 2.6 & 75.9 $\pm$
        2.7 & 79.8 $\pm$ 1.4 & 76.8 $\pm$ 2.1 & 76.4 $\pm$ 1.7 & 80.2 $\pm$
        1.4  \\
        \hline
        NN & 70.1 $\pm$ 3.1 & 72.8 $\pm$ 2.4
           & 82.3 $\pm$ 1.0 & 74.1 $\pm$ 1.6 & 75.8 $\pm$ 1.8 & 84.0 $\pm$ 1.5
        \\
        \hline
        NN MMD${}^*$ & 75.6 $\pm$ 2.9 & 78.4 $\pm$
           1.6 & 83.0 $\pm$ 1.2 & 77.9 $\pm$ 1.6 & 78.0 $\pm$ 1.9 & 84.7 $\pm$
        1.6  \\
        \hline
        NN MMD & \textbf{76.8 $\pm$ 2.0} & \textbf{79.1 $\pm$
          1.6} & \textbf{83.9 $\pm$ 1.0} & \textbf{78.3 $\pm$ 1.4} &
          \textbf{78.6 $\pm$ 2.6} & \textbf{85.2 $\pm$ 1.1}  \\
        \bottomrule
    \end{tabular}
    \caption{Domain adaptation results for product review sentiment
    classification task. NN MMD${}^*$: neural net with MMD trained and tested on word count
    instead of TF-IDF features.}
    \label{tbl:da_full}
\end{table}

The dataset contains 2000 product reviews in each of the 4 domains. Each
product review is represented as a bag of words and bigrams.  We preprocessed
the data and ignored all words and bigrams occurring less than 50 times across
the whole dataset. Then computed the new word-count vectors and TF-IDF vectors
for each product review and use these vectors as input representations of the
data.

To make the experiment results robust to sampling noise, we generated 10
random splits of the dataset, where each domain is split into 1500 examples
for training, 100 for validation and 400 for testing. For each domain
adaptation task from one source domain to a target domain, the training data
in the source domain is used as labeled source domain data, and the training
data without labels in the target domain is used as unlabeled target domain
data. The validation data in the source domain is used for early stopping in
neural network training, and the prediction accuracy on the test data from
target domain is used as the evaluation metric. For each of the methods we
considered in the experiments, hyper parameters are tuned to optimize the
average target domain prediction accuracy across all 10 random splits, and the
best average accuracy is reported, which is a setting similar to
cross-validation for domain adaptation tasks.

We used a fully connected neural network with two hidden layers, $128$ hidden
units on the first layer and $64$ hidden units on the second. All hidden units are rectified linear units (ReLU).
The MMD penalty is applied on the second hidden layer. We used Gaussian
kernels for MMD.  The final objective is composed of a classification
objective on the source domain and a MMD penalty for the source and target
domains. The model is trained using stochastic gradient descent, where the
initial learning rate is fixed and gradually adapted according to AdaGrad
\cite{duchi2011adaptive}. The hyperparameters of the model include the scale parameter in
Gaussian kernel and the weight for the MMD penalty. The learning rate,
momentum, weight-decay and dropout rate for neural network training are fixed for all
the experiments.

For TCA baseline, we tried both linear kernels and Gaussian RBF kernels, and
found that linear kernels actually works better, so the reported results are
all from linear kernel TCA models.  The projection matrix after kernel
transformation projects the examples down to 64 dimensions (same as the 2nd
hidden layer of the neural net above). Then a Gaussian kernel RBF SVM is
trained on the mean-std normalized projected features (we've tried linear SVMs as well but found
RBF SVMs work better).  We found the normalization step to be critical to the
performance of TCA as the scale of the features can differ by a few orders of
magnitudes.

Full results on all source-target pairs are shown in Table \ref{tbl:da_full}.
NN MMD with word count features are shown as ``NN MMD${}^*$''. Overall all
methods gets a significant boost from using TF-IDF features. But NN MMD method
is able to learn useful features for domain adaptation even with word count
features, and performs better than the baselines on most tasks.


\subsection{Relationship Between Contractive Auto-Encoders and MMD}
It is straightforward to show that the contractive auto-encoder can be written as an MMD penalty with a linear kernel. First take $e_i$ to be an elementary vector with a $1$ at index $i$ and $0$ everywhere else. We will take a Taylor expansion of a hidden unit $h_j(x)$ around $e_i$~\cite{rippel2014nested}:
\begin{align}
h_j(x + \epsilon e_i) &\approx h_j(x) + \epsilon e_i^\top \nabla h_j(x) + o(\epsilon^2), \\
\frac{h_j(x + \epsilon e_i) - h_j(x)}{\epsilon} &\approx e_i^\top \nabla h_j(x), \\
h_j(x + \epsilon e_i) - h_j(x) &\approx \epsilon \frac{\partial h_j(x)}{x_i}.
\end{align}
Squaring both sides and summing over each hidden dimension and data dimension recovers the contractive auto-encoder penalty.
\begin{align}
\sum_j \sum_i ( h_j(x + \epsilon e_i) - h_j(x) )^2 &\approx \epsilon^2 \sum_j \sum_i \left(\frac{\partial h_j(x)}{x_i} \right)^2.
\end{align}
The left hand side can be rewritten as an MMD penalty $||h(x) - \tilde{h}(x)||^2$, where $\tilde{h}(x) = [h_1(x + \epsilon e_1), h_2(x + \epsilon e_1), \ldots, h_K(x + \epsilon e_D)]$, assuming $K$ hidden units and $D$ data dimensions. Since there is no feature expansion, this is equivalent to using a linear kernel.

\subsection{Auto-Encoder Training Details}
We use a stochastic variant of the contraction penalty, where we sample $\tilde{h}(x)$ from a noise distribution. As in~\cite{vincent2010stacked}, we use Bernoulli noise where each data dimension is zeroed out with probability $p$, which is tuned along with the other hyperparameters. We use MMD with a Gaussian kernel $K(h(x),\tilde{h}(x)) = \exp(-\frac{1}{\sigma^2} ||h(x) - \tilde{h}(x)||^2)$. The networks each have one layer of $100$ sigmoidal hidden units and are trained using stochastic gradient descent with momentum.

\subsection{Auto-Encoder Weight Filters}
\begin{figure}[h]
    \centering
    \begin{subfigure}[b]{0.3\textwidth}
        \includegraphics[width=\textwidth]{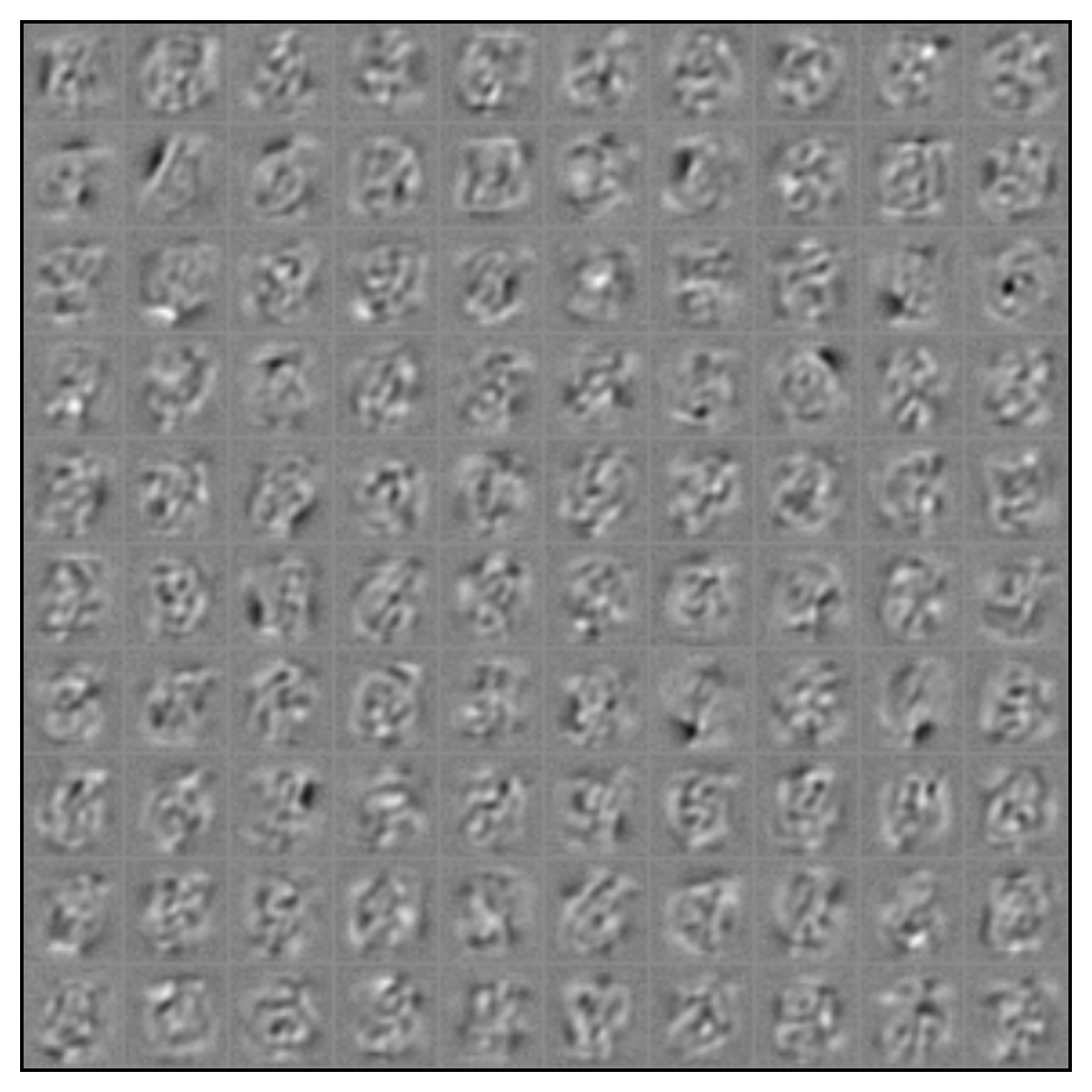}
        \caption{AE}
    \end{subfigure}
    \begin{subfigure}[b]{0.3\textwidth}
        \includegraphics[width=\textwidth]{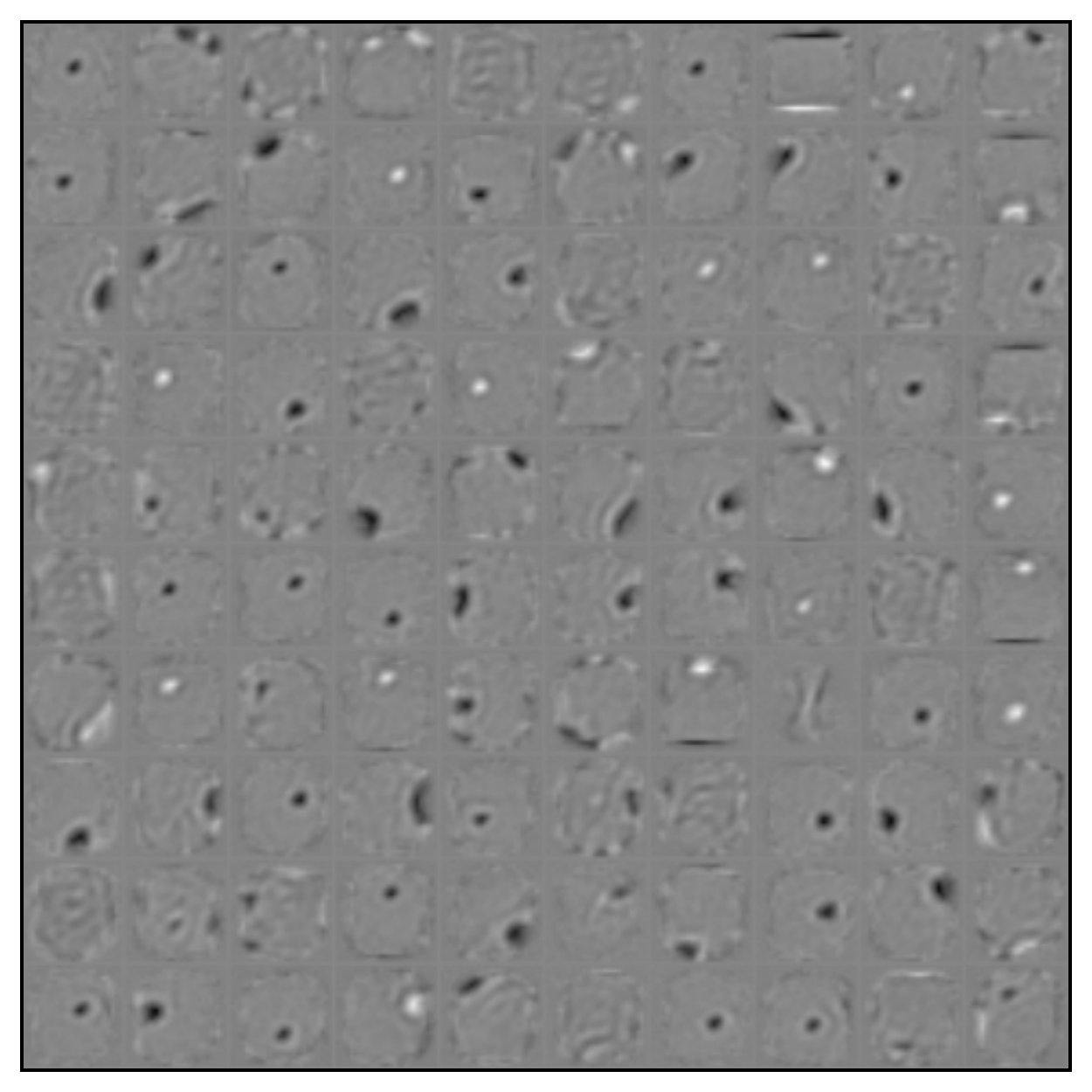}
        \caption{DAE}
    \end{subfigure}
    \begin{subfigure}[b]{0.3\textwidth}
        \includegraphics[width=\textwidth]{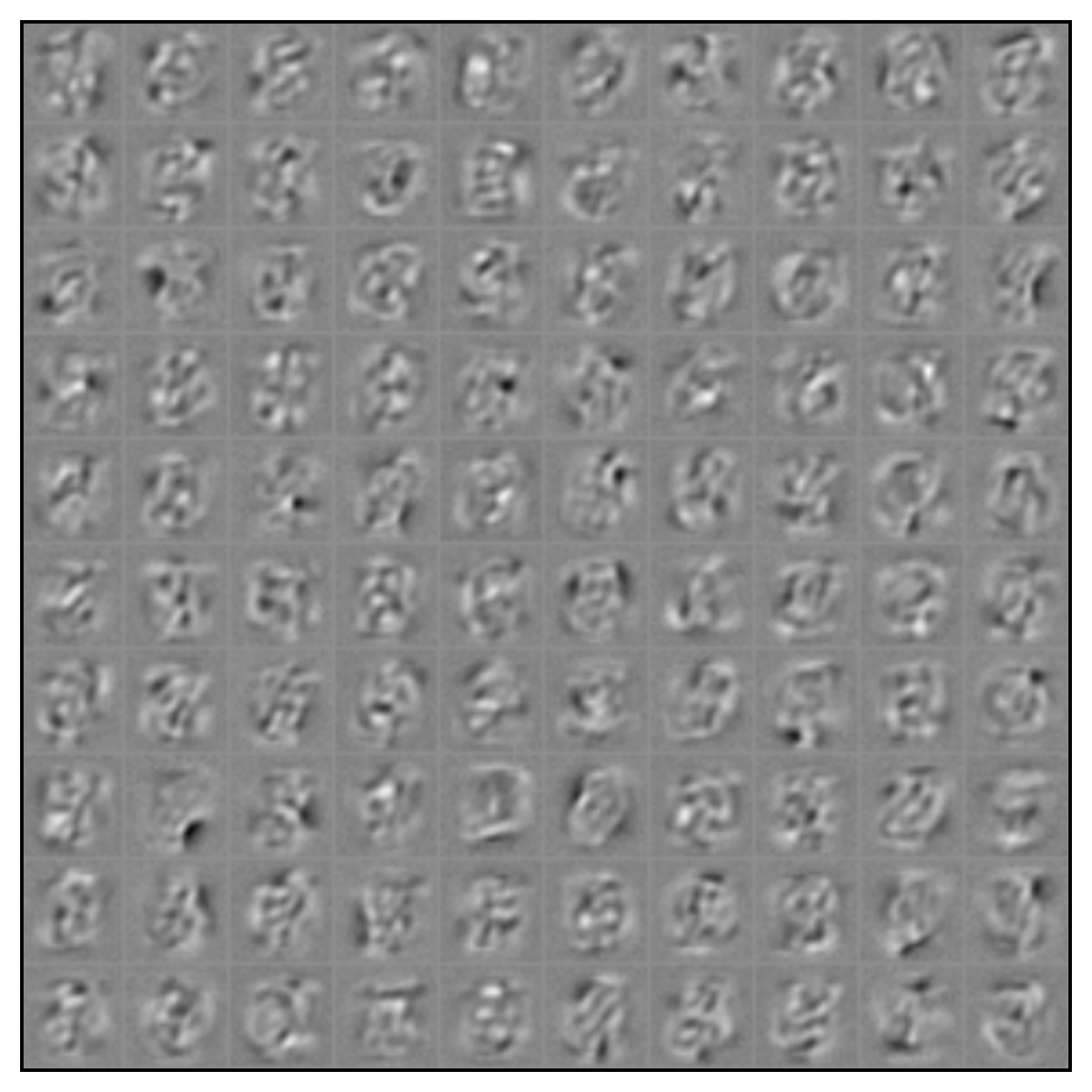}
        \caption{CAE}

    \end{subfigure}
    \begin{subfigure}[b]{0.3\textwidth}
        \includegraphics[width=\textwidth]{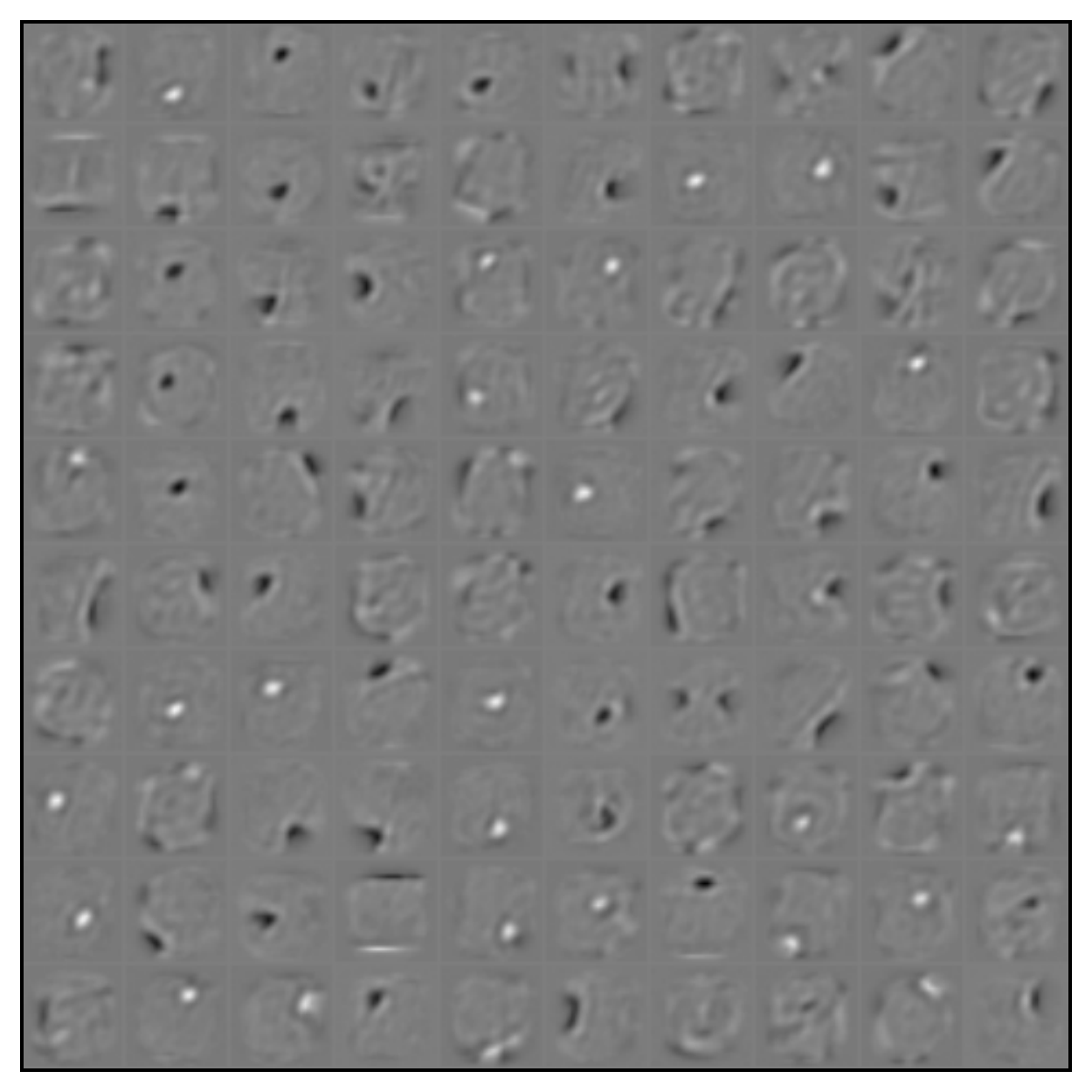}
        \caption{MMD}
    \end{subfigure}
    \begin{subfigure}[b]{0.3\textwidth}
        \includegraphics[width=\textwidth]{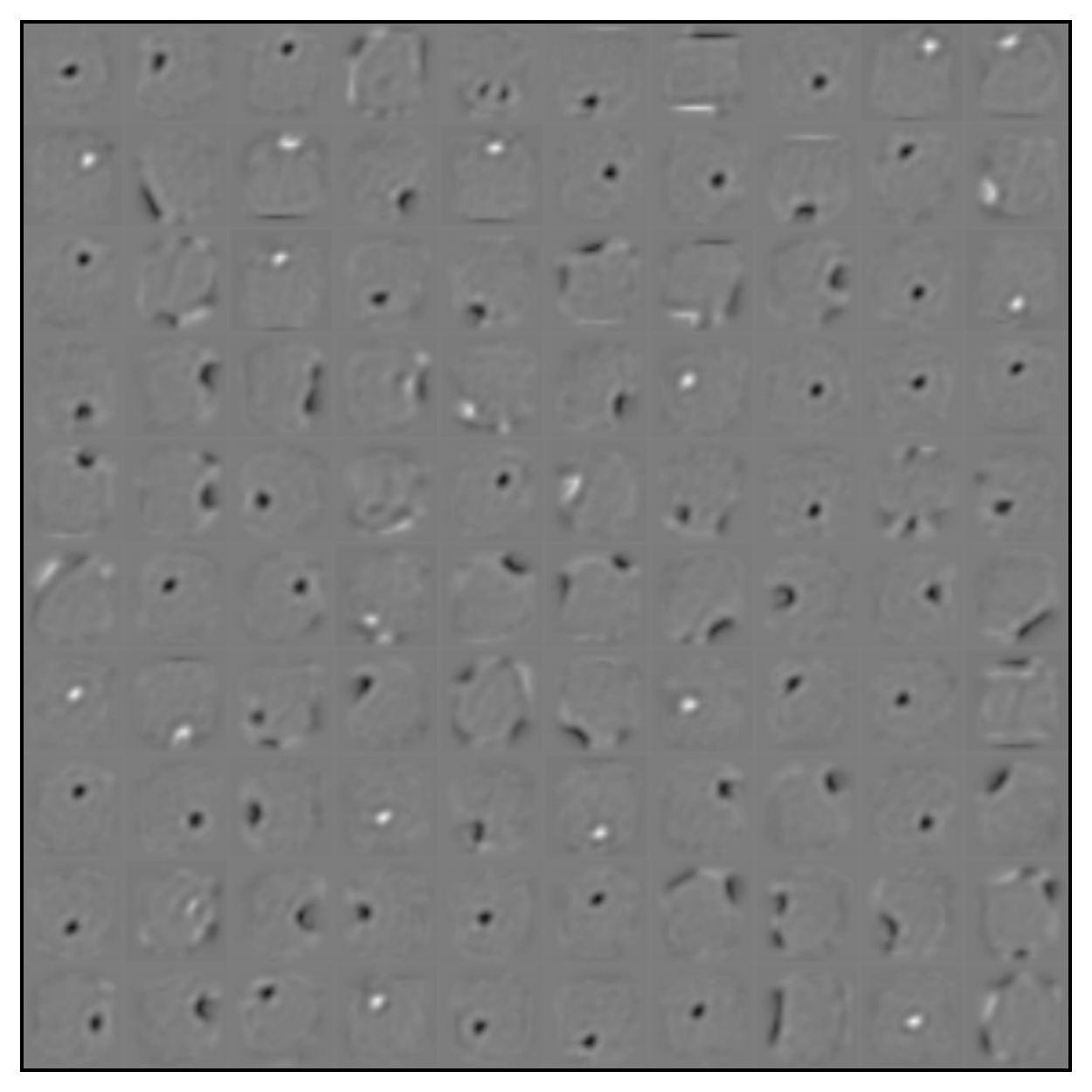}
        \caption{MMD+DAE}
    \end{subfigure}
    \caption{Visualization of the weight matrices for each variety of auto-encoder.}
    \label{fig:filters}
\end{figure}

Figure \ref{fig:filters} shows the weight filters, the weights from the each hidden unit to the data visualized as images. The MMD filters tend to be cleaner and more localized than the other variants.

\subsection{Training Details for the Generative Experiments}
We learn a generative deep model with 32 stochastic hidden
units with independent uniform prior distributions in $[-1,1]$, the
deterministic mapping is implemented by a feedforward network with two ReLU
layers with 64 and 128 units each, and then a final sigmoid layer of 784 units
(MNIST images are of size $28\times28=784$). We use a Gaussian kernel for the
MMD. For training, a set of new samples $h$ is generated from $p(h)$ after
every 200 updates to $f$.

\end{document}